\documentclass[11pt,a4paper]{article}
\usepackage[hyperref]{emnlp2020}
\usepackage{times}
\usepackage{latexsym}
\usepackage{amsmath}
\usepackage{graphicx}
\usepackage{cleveref}
\usepackage{multirow}
\usepackage{bm}
\crefname{section}{§}{§§}
\Crefname{section}{§}{§§}

\usepackage{xr-hyper}
\newcommand{\modelname}{FAST}
\usepackage{color}
\usepackage{microtype}

\aclfinalcopy 
\def\aclpaperid{1187} 

\title{Neural Deepfake Detection with Factual Structure of Text}

\author{Wanjun Zhong$^1$\thanks{\ \ \ Work done while this author was an intern at Microsoft Research.} , Duyu Tang$^2$, Zenan Xu$^1$$^*$, Ruize Wang$^3$, Nan Duan$^2$, Ming Zhou$^2$\\
	\bf Jiahai Wang$^1$ and Jian Yin$^1$\\
	$^1$ The School of Data and Computer Science, Sun Yat-sen University.\\
	Guangdong Key Laboratory of Big Data Analysis and Processing, Guangzhou, P.R.China\\
	$^2$ Microsoft Research  $^3$ Fudan University, Shanghai, P.R.China \\
	{\tt \{zhongwj25@mail2,xuzn@mail2\}.sysu.edu.cn}\\
	{\tt \{wangjiah@mail,issjyin@mail\}.sysu.edu.cn} \\
	{\tt \{dutang,nanduan,mingzhou\}@microsoft.com}\\ 
	{\tt rzwang18@fudan.edu.cn}\\
}

\date{}

\begin{document}
	\maketitle
	\begin{abstract}
		Deepfake detection, the task of automatically discriminating machine-generated text, is increasingly critical with recent advances in natural language generative models. Existing approaches to deepfake detection typically represent documents with coarse-grained representations. However, they struggle to capture factual structures of documents, which is a discriminative factor between machine-generated and human-written text according to our statistical analysis. To address this, we propose a graph-based model that utilizes the factual structure of a document for deepfake detection of text. Our approach represents the factual structure of a given document as an entity graph, which is further utilized to learn sentence representations with a graph neural network. Sentence representations are then composed to a document representation for making predictions, where consistent relations between neighboring sentences are sequentially modeled. Results of experiments on two public deepfake datasets show that our approach significantly improves strong base models built with RoBERTa. Model analysis further indicates that our model can distinguish the difference in the factual structure between machine-generated text and human-written text.


	\end{abstract}
	
	\vspace{0.0001cm}
	\section{Introduction}
	
	Nowadays, unprecedented amounts of online misinformation (e.g., fake news and rumors) spread through the internet, which may misinform people's opinions of essential social events \cite{faris2017partisanship,thorne2018fever,goodrich2019assessing,kryscinski2019evaluating}. 
	Recent advances in neural generative models, such as GPT-2 \cite{radford2019language}, make the situation even severer as their ability to generate fluent and coherent text may enable adversaries to produce fake news.
	In this work, we study deepfake detection of text, to automatically discriminate machine-generated text from human-written text. 
	\par
	Previous works on deepfake detection of text are dominated by neural document classification models \cite{bakhtin2019real,zellers2019defending,wang2019weak,vijayaraghavan2020fake}.
	They typically tackle the problem with coarse-grained document-level evidence such as dense vectors learned by neural encoder and traditional features (e.g.,  TF-IDF, word counts).
	However, these coarse-grained models struggle to capture the fine-grained factual structure of the text. 
	We define the factual structure as a graph containing entities mentioned in the text and the semantically relevant relations among them.
	\begin{figure}[t]
		\centering
		\includegraphics[width=0.47\textwidth]{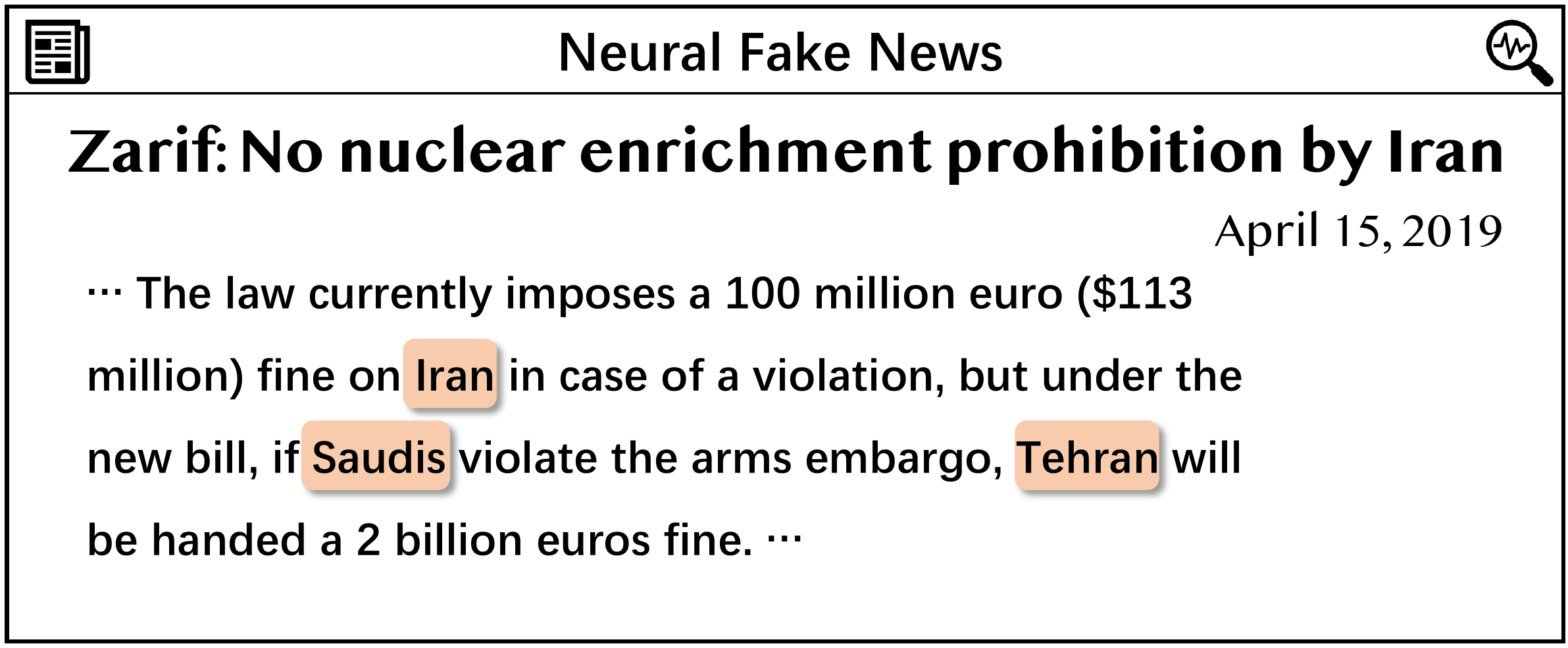}
		\caption{An example of machine-generated fake news. 
			We can observe that the factual structure of entities extracted by named entity recognition is inconsistent.
		}
		\label{fig:example}
	\end{figure} 
	As shown in the motivating example in Figure \ref{fig:example}, even though machine-generated text seems coherent, its factual structure is inconsistent. 
	Our statistical analysis further reveals the difference in the factual structure between human-written and machine-generated text (detailed in Section \ref{sec:data-analysis}). 
	Thus, 
	modeling factual structures is essential for detecting machine-generated text.
	\par
	Based on the aforementioned analysis, we propose \textbf{\modelname}, a graph-based reasoning approach utilizing \textbf{FA}ctual \textbf{S}tructure of \textbf{T}ext for deepfake detection.
	With a given document, we 
	represent its factual structure as a graph, where nodes are automatically extracted by named entity recognition. 
	Node representations are calculated not only with the internal factual structure of a document via a graph convolution network, but also with external knowledge from entity representations pre-trained on Wikipedia. 
	These node representations are fed to
	produce sentence representations which, together with the coherence of continuous sentences, are further used to compose a document representation for making the final prediction. 
	\par
	We conduct experiments on a news-style dataset and a webtext-style dataset, with negative instances generated by GROVER \cite{zellers2019defending} and  GPT-2 \cite{radford2019language} respectively. 
	Experiments show that our method significantly outperforms strong transformer-based baselines on both datasets.
	Model analysis further indicates that our model can distinguish the difference in the factual structure between machine-generated text and human-written text. 
	The contributions are summarized as follows:
	\begin{itemize}
		\item We propose a graph-based approach, which models the fine-grained factual structure of a document
		for deepfake detection of text.
		\item We statistically show that machine-generated text differs from human-written text in terms of the factual structures, and  
		injecting factual structures boosts detection accuracy. 
		\item Results of experiments on news-style and webtext-style datasets verify that our approach achieves improved accuracy compared to strong transformer-based pre-trained models.
	\end{itemize}
	
	\section{Task Definition}
	We study the task of deepfake detection of text in this paper. 
	This task discriminates machine-generated text from human-written text, which can be viewed as a binary classification problem. 
	We conduct our experiments on two datasets with different styles: a news-style dataset with fake text generated by GROVER \cite{zellers2019defending} and a large-scale webtext-style dataset with fake text generated by GPT-2 \cite{radford2019language}. The news-style dataset consists of 25,000 labeled documents, and the webtext-style dataset consists of 520,000 labeled documents.
	With a given document, systems are required to perform reasoning about the content of the document and assess whether it is ``human-written" or ``machine-generated".
	\section{Factual Consistency Verification}
	\label{sec:data-analysis}
	In this part, we conduct a statistical analysis to reveal the difference in the factual structure between human-written and machine-generated text. Specifically, we study the difference in factual structures from a consistency perspective and analyze entity-level and sentence-level consistency.
	
	Through data observation, we find that human-written text tends to repeatedly mention the same entity in continuous sentences, while machine-written continuous sentences are more likely to mention irrelevant entities. 
	Therefore, we define \textbf{entity consistency count (ECC)} of a document as the number of entities that are repeatedly mentioned in the next $w$ sentences, where $w$ is the sentence window size. 
	\textbf{Sentence consistency count (SCC)} of a document is defined as the number of sentences that mention the same entities with the next $w$ sentences.
	For instance, if entities mentioned in three continuous sentences are ``\textit{A and B; A; B}" and $w=2$, then $ECC=2$ because two entities $A$ and $B$ are repeatedly mentioned in the next 2 sentences. $SCC=1$ because only the first sentence has entities mentioned in the next 2 sentences.
	We use all 5,000 pairs of human-written and machine-generated documents from the news-style dataset and each pair of documents share the same metadata (e.g., title) for statistical analysis. We plot the kernel density distribution of these two types of consistency count with sentence window size $w=\{1,2\}$.
	
	\begin{figure}[h]
		\centering
		\includegraphics[width=0.48\textwidth]{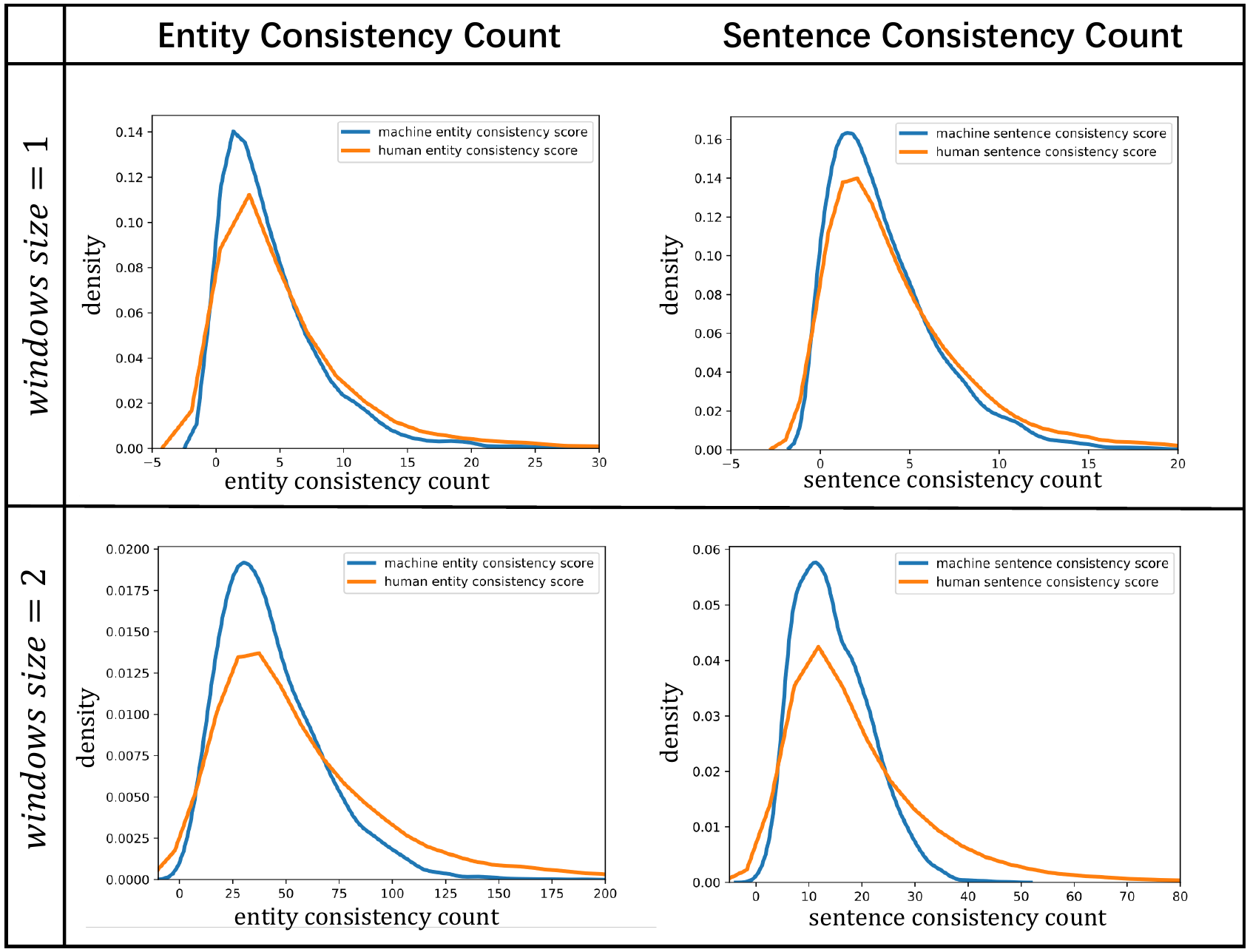}
		\caption{Statistical analysis about entity-level and sentence-level consistency. 
			Orange curve and blue curve indicate kernel density estimation curve for human-written document and machine-generated document respectively.
			X-axis indicates the value of consistency count and y-axis indicates probability density.
		}
		\label{fig:data-analysis}
	\end{figure} 
	\begin{figure*}[t]
		\centering
		\includegraphics[width=0.98\linewidth]{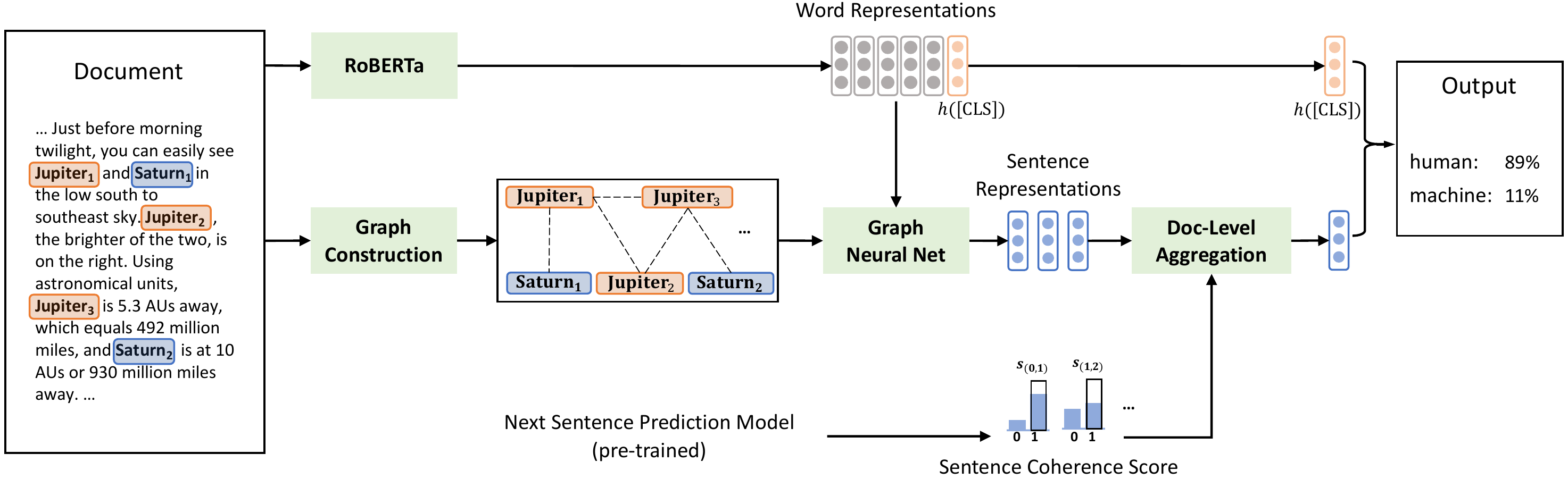}
		\caption{An overview of our approach. Taking a document as the input, we first calculate contextual word representations via RoBERTa (\cref{sec:word-rep}) and represent the factual structure as a graph (\cref{sec:graph-construct}). After that, we employ graph neural network to learn sentence representations (\cref{sec:gcn}). Then, sentence representations are composed to a document representation considering coherence of continuous sentences before making the final prediction (\cref{sec:doc-rep}).
		}
		\label{fig:pipeline}
	\end{figure*}
	 As shown in Figure \ref{fig:data-analysis}, human-written documents are more likely to 
	have higher entity-level and sentence-level consistency count. 
	This analysis indicates that human-written and machine-generated text are different in the factual structure, thus modeling consistency of factual structures is essential in discriminating them.
	
	\section{Methodology}
	
	In this section, we present our graph-based reasoning approach, which considers factual structures of documents, which is used to guide the reasoning process for the final prediction.
	\par
	Figure \ref{fig:pipeline} gives a high-level overview of our approach. 
	With a document given as the input, our system begins by calculating the contextual word representations by RoBERTa (\cref{sec:word-rep}). 
	Then, we build a graph for capturing the internal factual structure of the whole document  (\cref{sec:graph-construct}).
	With the constructed graph, we initialize node representations utilizing internal and external factual knowledge and propagate and aggregate information by a graph neural network to learn graph-enhanced sentence representations (\cref{sec:gcn}). 
	Then, to model the consistent relations of continuous sentences and compose a document representation for making the final prediction, we employ a sequential model with help of coherence scores from a pre-trained next sentence prediction (NSP) model (\cref{sec:sentence-coherence}).
	
	\par
	\subsection{Word Representation}
	\label{sec:word-rep}
	In this part, we present how to calculate contextual word representations by a transformer-based model. In pratice, we employ RoBERTa \cite{liu2019roberta}.
	\par
	Taking a document $d$ as the input, we employ RoBERTa to learn contextual semantic representations for words \footnote{In practice, ``words" may indicate tokens or token-pieces, we use ``words" for a better illustration here. }. RoBERTa encoder $\mathcal{B}$ maps document $\bm{x}$ with length $|\bm{x}|$ into a sequence of following hidden vectors.
	\begin{equation}
	\bm{h}(\bm{x}) = [\bm{h}(\bm{x})_1,\bm{h}(\bm{x})_2,\cdots,\bm{h}(\bm{x})_{|\bm{x}|}]
	\end{equation}
	where each $\bm{h}(\bm{x})_i$ indicates the contextual representation of word $i$
	\subsection{Graph Construction}
	\label{sec:graph-construct}
	In this part, we present how to construct a graph to reveal the internal factual structure of a document. 
	In practice, we observe that selecting entities, the core participants of events, as arguments to construct the graph leads to less noise to the representation of the factual structure. 
	Therefore, we employ a named entity recognition (NER) model to parse entities mentioned in each sentence. 
	Specifically, taking a document as the input, we construct a graph in the following steps.
	\begin{itemize}
		\item We parse each sentence to a set of entities with an off-the-shelf NER toolkit built by AllenNLP \footnote{\url{https://demo.allennlp.org/named-entity-recognition}}, which is an implementation of \citet{peters2017semi}. Each entity is regarded as a node in the graph.
		\item We establish links between inner-sentence and inter-sentence entity node pairs to capture the structural relevance.
		We add inner-sentence edges to entity pairs in the same sentence for they are naturally relevant to each other. 
		Moreover, we add inter-sentence edges to literally similar inter-sentence entity pairs for they are likely to be the same entity.
	\end{itemize}
	After this process, the graph reveals the fine-grained factual structure of a document.
	
	\subsection{Graph Neural Network}
	\label{sec:gcn}
	In this part, we introduce how to initialize node representations and exploit factual structure utilizing multi-layer graph convolution network (GCN) to propagate and aggregate information and finally produce sentence representations. 
	\subsubsection{Node Representation Initialization}
	We initialize node representations with contextual word representations learnt by RoBERTa and 
	external entity representations pre-trained on Wikipedia.
	\paragraph{Contextual Representation}
	Since each entity node is naturally a span of words mentioned in the document, we calculate the contextual representation of each node by the contextual words representations $\bm{h}(\bm{x})$. Supposing an entity $e$ consists of $n$ words, then the contextual representation $\bm{\varepsilon}_\mathcal{B}$ is calculated with the following formula:
	\begin{equation}
	\bm{\varepsilon}_\mathcal{B} = ReLU(\bm{W}_{\mathcal{B}} \frac{1}{n} \sum_{i=0}^{n} \bm{h}(\bm{x})_{e^i})
	\end{equation}
	where $\bm{W}_{\mathcal{B}} $ is a weight metric, $e^i$ is the absolute position in the document of the $i^{th}$ word in the span of entity $e$, and $ReLU$ is an activation function.
	\paragraph{Wikipedia-based Entity Representation}
	To model external factual knowledge about entities in the knowledge base, 
	we further represent entity $e$ with a projected wikipedia2vec entity representation  \cite{yamadawikipedia2vec}, which embeds words and entities on Wikipedia pages in a common space.
	The Wikipedia-based entity representation $\bm{\varepsilon}_w$ is :
	\begin{equation}
	\bm{\varepsilon}_w = ReLU(\bm{W}_w\bm{v}_e)
	\end{equation}
	where $\bm{v}_e$ is the wikipedia2vec representation of entity $e$ and $\bm{W}_w$ is a weight metric.
	\paragraph{}
	The initial representation $\bm{H}^{(0)}_e \in \bm{R}^d$ of entity node $e$ is the concatenation of contextual representation $\bm{\varepsilon}_\mathcal{B}$ and Wikipedia-based entity representation $\bm{\varepsilon}_w$, with dimension $d$.
	
	\subsubsection{Multi-layer GCN}
	In order to propagate and aggregate information through multihop neighbouring nodes, we employ multi-layer Graph Convolution Network (GCN) \cite{kipf2016semi}.
	
	Formally, we denote the constructed graph as $G$ and representation of all nodes as $\bm{H} \in \bm{R}^{N \times d}$, where $N$ denote the number of nodes. Each row $\bm{H}_e \in \bm{R}^d$ in $\bm{H}$ indicates a representation of node $e$. We denote the adjacency matrix of graph $G$ as $\bm{A}$ and degree matrix as $\bm{D}$.  We further calculate $\widetilde{\bm{A}}=\bm{D}^{-\frac{1}{2}}\bm{AD}^{-\frac{1}{2}}$.
	Then, the formula of multi-layer GCN is described as follows:
	\begin{equation}
	\bm{H}^{(i+1)}_e=\sigma(\widetilde{\bm{A}}\bm{H}_e^{(i)}\bm{W}_i)
	\end{equation}
	where $\bm{H}^{(i)}_e$ denotes the representation of node $e$ calculated by $i^{th}$ layer of GCNs, $\bm{W}_i$ is the weight matrix of layer $i$. $\sigma$ is an activation function. Specially, $\bm{H}^{(0)}_e$ is the initialized node representations. 
	
	Finally, through $m$ layers of GCN, we obtain the graph-enhanced node representations based on the structure of the factual graph.
	\subsubsection{Sentence Representation}
	According to compositionality, we believe that global representation should come from partial representations.
	Therefore, we calculate sentence-level representations based on graph-enhanced node representations. Supposing sentence $i$ has $N_i$ corresponding entities, we calculate the representation $\bm{y}_i$ of sentence $i$ as follows:
	\begin{equation}
	\bm{y}_i = \frac{1}{N_i} \sum_{j=0}^{N_i}\sigma(\bm{W_s} \bm{H}_{(i,j)}+\bm{b}_s)
	\end{equation}
	where $\sigma$ is an activation function, $\bm{W_s}$ is a weight matrix, $\bm{b}_s$ is a bias vector and $\bm{H}_{(i,j)}$ indicates the representation of $j^{th}$ node in sentence $i$. The compositionality can also be implemented in other ways, which we leave to future work.

	\subsection{Aggregation to Document Representation}
	\label{sec:doc-rep}
	In this part, we present how to compose a document representation for the final prediction utilizing graph-enhanced sentence representations and coherence score calculated by a pre-trained next sentence prediction (NSP) model.
	\paragraph{Coherence Tracking LSTM}
	With graph-enhanced sentence representations given as the input, the factual consistency of continuous sentences is automatically modeled by a sequential model.
	Specifically, We employ LSTM to track the consistent relations and produce representations $\widetilde{\bm{y}_i}$ for sentence $i$
	\begin{equation}
	\widetilde{\bm{y}_i} = LSTM([\bm{y}_i])
	\end{equation}
	
	\paragraph{Next Sentence Prediction Model}
	In order to further model contextual coherence of neighbouring sentence pairs as an additional information, we pre-train an NSP model to calculate the contextual coherence score for each neighbouring sentence pair. We employ RoBERTa \cite{liu2019roberta} as the backbone, which receives pairs of sentences as the input and assesses whether the second sentence is a subsequent sentence of the first. Further training details are explained in Appendix \ref{appendix:nsp}.
	The outputs $\bm{S}$  are described as follows. 
	\begin{equation}
	\bm{S} = [S_{(0,1)},...,S_{(s-1,s)}]
	\end{equation}
	where $s+1$ is the number of sentences in document $\bm{x}$ and each $S_{(i-1,i)}$ is the positive probability score for sentence pair ($i-1$, $i$), which indicates how likely it is that sentence $i$ is a subsequent sentence of sentence $i-1$.

	\paragraph{Prediction with NSP Score}
	We generate a document-level representation by composing sentence representations before making the final prediction.
	To achieve this, we take NSP scores as weights and calculate the weighted sum of representations of sentence pairs with the assumption that sentence pairs with higher contextual coherence score should also carry more importance in making the final prediction.  The final document representation $\bm{D}$ is calculated as follows.
	\begin{equation}
	\bm{D} =   \sum_{j=1}^{s} S_{(j-1,j)} * [\widetilde{\bm{y}}_{j-1}, \widetilde{\bm{y}_j}]
	\end{equation}
	Finally, we make the final prediction by feeding the combination of $\bm{D}$ and the last hidden vector $\bm{h}([CLS])$ from RoBERTa through an classification layer. 
	The goal of this operation is to maintain the complete contextual semantic meaning of the whole document because some linguistic clues are left out during graph construction.
	\label{sec:sentence-coherence}
	
	\section{Experiment}
	\subsection{Experiment Settings}

	In this paper, we evaluate our system on the following two datasets: 
	
	\begin{itemize}
		\item News-style GROVER-generated dataset provided by \citet{zellers2019defending}. The human-written instances are collected from RealNews, and machine-generated instances are generated by GROVER-Mega, a large state-of-the-art transformer-based generative model developed for neural fake news. We largely follow the experimental settings as described by \citet{zellers2019defending} and adopt two evaluation metrics: \textbf{paired accuracy} and \textbf{unpaired accuracy}. In the paired setting, the system is given human-written news and machine-generated news with the same meta-data. The system needs to assign higher machine probability to the machine-generated news than the human-written one.
		In the unpaired setting, the system is provided with single news document and states whether the document is human-written or machine-generated. 
		\item Webtext-style GPT2-generated dataset provided by OpenAI\footnote{\url{https://github.com/openai/gpt-2-output-dataset}}. The human-written instances are collected from WebText. Machine-generated instances are generated by GPT-2 XL-1542M \cite{radford2019language}, a powerful transformer-based generative model trained on a corpus collected from popular webpages. For this dataset, we adopt binary classification accuracy as the evaluation metric.
	\end{itemize}
	We set nucleus sampling  with $p=0.96$ as the sampling strategy of generator for both datasets, which leads to better generated text quality \cite{zellers2019defending,ippolito2019human}. 
	The statistics of the two datasets are shown in the Table \ref{tab:data}.
	\begin{table}[h]
		\scalebox{0.83}{
			\begin{tabular}{lcccc}
				\hline
				Dataset               & Train   & Valid  & \multicolumn{2}{c}{Test Set} \\
				&         &        & Unpaired       & Paired      \\ \hline
				News-style    & 10,000  & 3,000  & 8,000          & 8,000       \\
				Webtext-style & 500,000 & 10,000 & 10,000         & -           \\ \hline
		\end{tabular}}
		\caption{Statistics of news-style and webtext-style datasets.}
		\label{tab:data}
	\end{table}
	
	Furthermore, we adopt RoBERTa-Base \cite{liu2019roberta} as the direct baseline for our experiments because RoBERTa achieves state-of-the-art performance on several benchmark NLP tasks. 
	The hyper-parameters and training details of our model are described in Appendix \ref{appendix:fast}.
	\subsection{Model Comparison}

	\paragraph{Baseline Settings} 
	\begin{table}[t]
		\scalebox{0.83}{
			\begin{tabular}{llcc}
				\hline
				Size               & Model        & Unpaired Acc & Paired Acc \\ \hline
				& Chance       & 50.0\%         & 50.0\%       \\ \hline
				\multirow{3}{*}{355M} & GROVER-Large & 80.8\%       & 89.0\%     \\
				& BERT-Large   & 73.1\%       & 84.1\%     \\
				& GPT2         & 70.1\%       & 78.8\%     \\ \hline
				\multirow{6}{*}{124M} & GROVER-Base  & 70.1\%       & 77.5\%     \\
				& BERT-Base    & 67.2\%       & 80.0\%     \\
				& GPT2         & 66.2\%       & 72.5\%     \\
				& XLNet        & 77.1\%      & 88.6\%    \\
				& RoBERTa      & 80.7\%      & 89.2\%    \\ \cline{2-4} 
				& \modelname    & \textbf{84.9\%}      & \textbf{93.5\%}    \\ \hline
		\end{tabular}}
		\caption{Performance on the test set of news-style dataset in terms of unpaired and paired accuracy. Our model is abbreviated as \modelname.  Size indicates approximate model size. 
			The performance of GROVER, BERT, and GPT2 are reported by \citet{zellers2019defending}}
		\label{tab:grover-model-cmp}
	\end{table}
	We compare our system with transformer-based baselines for DeepFake detection, including three powerful transformer-based pre-trained models: \textbf{BERT} \cite{devlin2018bert}, \textbf{XLNet} \cite{yang2019xlnet} and \textbf{RoBERTa} \cite{liu2019roberta}, which are large bidirectional transformers achieving strong performance on multiple benchmark NLP tasks. These baselines add a simple classification layer on top of them and are fine-tuned with standard cross-entropy loss on the binary classification. 
	\par 
	For the news-style dataset, we further compare our model with \textbf{GPT-2} \cite{radford2019language} and \textbf{GROVER} \cite{zellers2019defending}. The GROVER-based discriminator is a fine-tuned version of generator GROVER, which has three model sizes: GROVER-Base (124 million parameters), GROVER-Large (335 million parameters), and GROVER-Mega (1.5 billion parameters). 
	Our model is not comparable with GROVER-Mega for the following reasons. 
	Firstly, GROVER-Mega is the fake news generator, and it has a strong inductive bias (e.g., data distribution and sampling strategy) as the discriminator \cite{zellers2019defending}. 
	Secondly, GROVER-Mega has a much larger model size (1.5 billion parameters) than our model. 
	
	For the webtext-style dataset, we compare with the baselines we trained with the same hyper-parameters. 
	We don't compare with GPT-2 because it's the generator for machine-generated text.
	
	\paragraph{Results and Analysis}

	In Table \ref{tab:grover-model-cmp}, we compare our model with baselines on the test set of news-style dataset with negative instances generated by GROVER-Mega. As shown in the table, our model significantly outperforms our direct baseline RoBERTa with 4.2\% improvements on unpaired accuracy and 4.3\% improvements on paired accuracy. Our model also significantly outperforms GROVER-Large and other strong transformer-based baselines (i.e., GPT2, BERT, XLNet). 

	In Table \ref{tab:gpt2-model-cmp}, we compare our model with baselines on the development set and the test set of webtext-style dataset. Our model significantly outperforms strongest transformer-based baseline RoBERTa by 2.64\% on the development set and 3.07\% on the test set of webtext-style GPT2-generated dataset. 
	
	This observation indicates that modeling fine-grained factual structures empower our system to discriminate the difference between human-written text and machine-generated text.
	\begin{figure*}[t]
		\centering
		\includegraphics[width=0.95\linewidth]{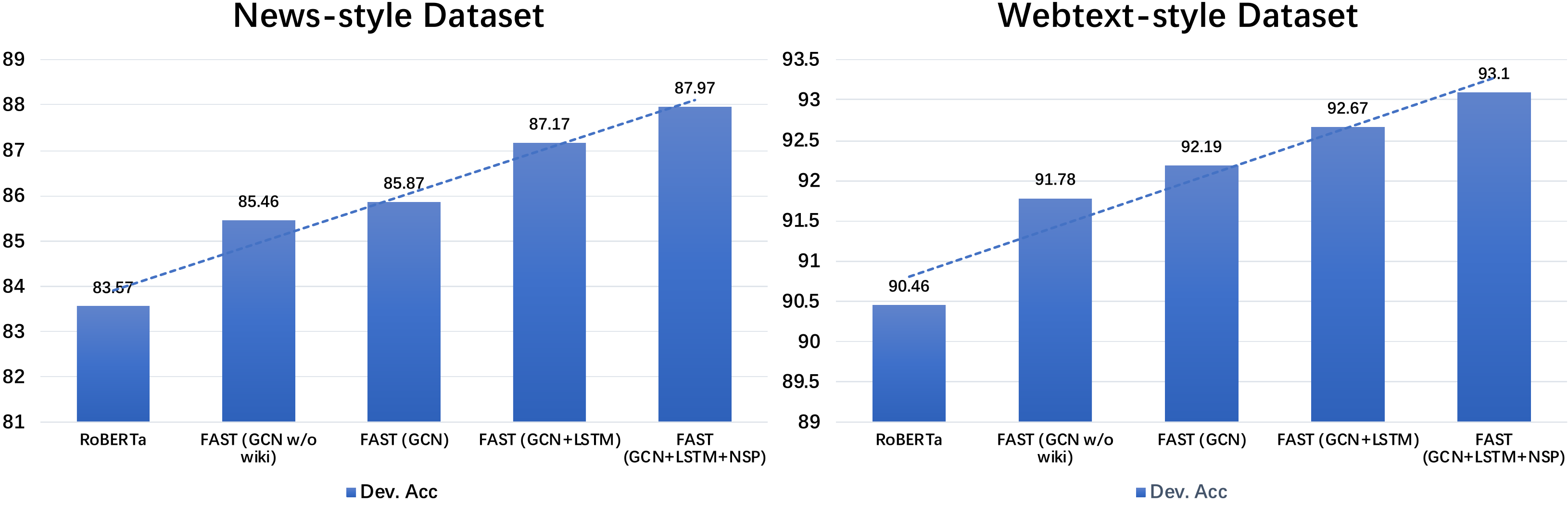}
		\caption{Ablation studies on the the development set of the two datasets in terms of binary classification accuracy.}
		\label{fig:ablation-study}
	\end{figure*}

	\subsection{Ablation Study}
	\begin{table}[t]
		\centering
		\begin{tabular}{lcc}
			\hline
			Model                                                                     & Development Acc & Test Acc \\ \hline
			Random                                                                    & 50.00\%        & 50.00\%        \\ \hline
			BERT                                                               &      85.32\%      & 85.10\%          \\
			XLNet                                                              &      88.79\%         &     88.56\%          \\
			RoBERTa                                                              & 90.46\%       & 90.10\%       \\ \hline
			\modelname & \textbf{93.10\%} & \textbf{93.17\%} \\ \hline
		\end{tabular}
		\caption{Performance on the development and test set of webtext-style dataset in terms of binary classification accuracy. Our model is abbreviated as \modelname.}
		\label{tab:gpt2-model-cmp}
	\end{table}
	Moreover, we also conduct ablation studies to evaluate the impact of each component by conducting experiments about direct baseline RoBERTa-Base and four variants of our full model. 
	\begin{itemize}
		\item \textbf{RoBERTa-Base} is our direct baseline without considering any structural information.
		\item \textbf{\modelname\ (GCN)} calculate a global document representation by averaging node representations after representation learning by GCN.
		\item \textbf{\modelname\ (GCN w/o wiki)} The node representations eliminate entity representations from wikipedia2vec and the rest are the same as \modelname\ (GCN).
		
		\item \textbf{\modelname\ (GCN + LSTM)} takes the final hidden state from coherence tracking LSTM (\cref{sec:doc-rep}) as the final document-level representation.
		\item \textbf{\modelname\ (GCN + LSTM + NSP)} is the full model introduced in this paper.
	\end{itemize}
	As shown in Figure \ref{fig:ablation-study}, adding GCN improve performance on the development the set of news-style dataset and webtext-style dataset. 
	This verifies that incorporating fine-grained structural information is beneficial for detecting generated text. 
	Eliminating wikipedia-based entity representation from \modelname\ (GCN) drops performance, which indicates that incorporating external knowledge is also beneficial.
	Moreover, incorporating coherence tracking LSTM brings further improvement on two datasets, which indicates that modeling consistency of factual structure of continuous sentences is better than simply using global structural information of the document, like the setting in \modelname\ (GCN). Lastly, results also show that incorporating semantic coherence score of pre-trained NSP model is beneficial for discriminating generated text.

	\subsection{Case Study}
	\begin{figure*}[t]
		\centering
		\includegraphics[width=0.95\linewidth]{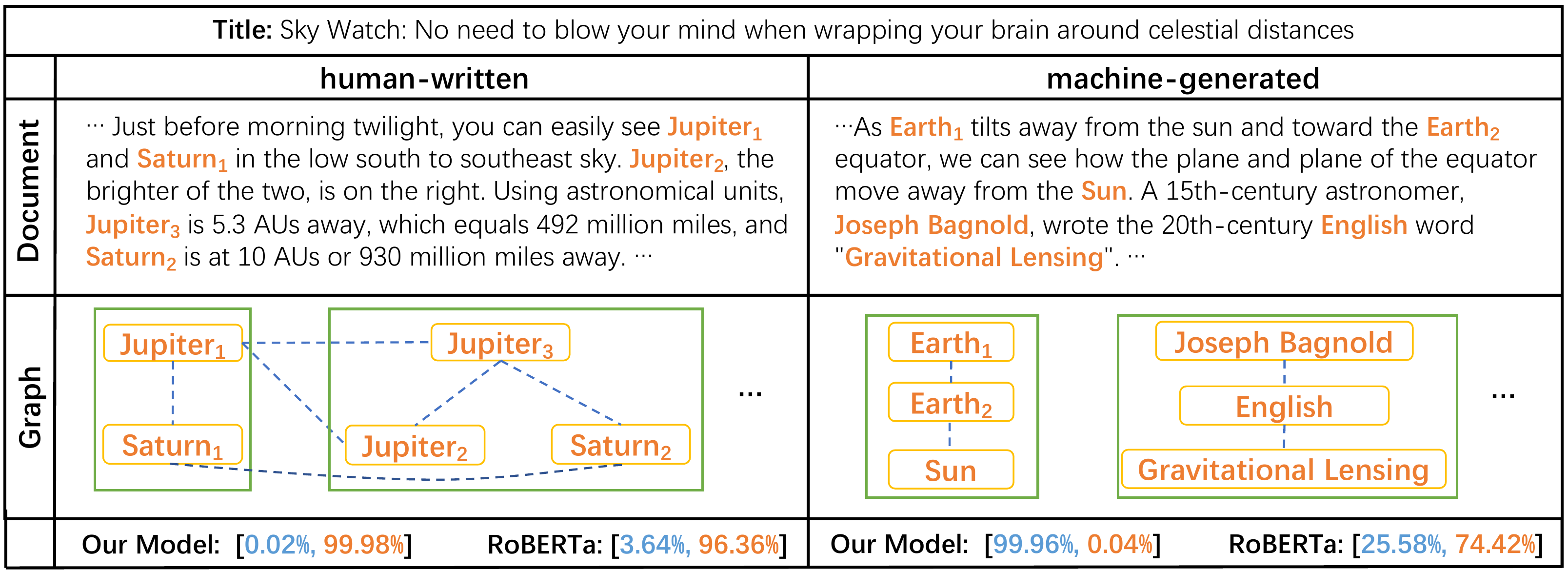}
		\caption{A case study of our approach. Continuous words in orange indicate a entity node extracted by our system. Each green solid box indicates a sub-graph  corresponding to a sentence, 
			and a blue dashed line indicates an edge between semantically relevant entity pairs. Numbers in orange and blue indicate probability for the human-written document and the machine-generated document respectively.}
		\label{fig:case-study}
	\end{figure*}
	As shown in Figure \ref{fig:case-study}, we conduct a case study by giving an example.
	This example shows human-written news and machine-generated news with the same metadata (i.e., title). The veracity of both documents are correctly predicted by our model.
	With the given document, our system constructs a factual graph and makes the correct predictions by reasoning over the constructed graph. 
	We can observe that although the continuous sentences in the machine-generated news look coherent, their factual structure is not consistent as they describe events about irrelevant entities.
	Instead, the human-written news has a more consistent factual structure.
	However, without utilizing factual structure information, RoBERTa fails to discriminate between these two articles.
	This observation reflects that our model can distinguish the difference in the factual consistency of machine-generated text and human-written text. 
	\subsection{Error Analysis}
	To explore further directions for future studies, we randomly select 200 instances and manually summarize representative error types.
	
	The \textbf{primary} type of errors is those caused by failing to extract core entities of sentences. The quality of a constructed graph is somehow limited by the performance of the NER model. 
	This limitation leaves further exploratory space for extraction of internal factual structure. 
	The \textbf{second} type of errors is caused by the weakness in the mathematical calculation of the model. For instance, a document describes that \textit{``a smaller \textcolor[rgb]{0,0,1}{\$5} million one-off was seized in 2016 and the National Bank of Antigua and Barbuda reclaimed \textcolor[rgb]{0,0,1}{\$30} million stolen in the 2015 heist last year.  \textcolor[rgb]{0,0,1}{\$100} million, it was a massive amount. But now we are talking of \textcolor[rgb]{0,0,1}{\$50} million, this is extremely conservative... "}. 
	Humans can easily observe that the mentioned numbers are highly inconsistent in the generated text. A machine struggles to discern that. 
	This error type calls for the development of a machine's mathematical calculation abilities. 
	The \textbf{third} error type is caused by failing to model commonsense knowledge. For example, a famous generated document mentioned \textit{``In a shocking finding, scientist discovered a herd of \textcolor[rgb]{0,0,1}{unicorns} living in a remote, previously unexplored valley, in the Andes Mountains. ... These \textcolor[rgb]{0,0,1}{four-horned, silver-white unicorns} were previously unknown to science."}. Although the text looks coherent, it is still problematic in terms of commonsense knowledge that ``unicorn has only one horn". This leaves space for further research on exploring commonsense knowledge in deepfake detection.

	\section{Related Work}
	Recently, fake news detection has attracted growing interest due to the unprecedented amount of fake contents propagating through the internet \cite{vosoughi2018spread}. Spreading of fake news arises public concerns \cite{cooke2018fake} as it may influence essential public events like politic elections \cite{allcott2017social}. Online reviews can also be generated by machines, and can even be as fluent as human-written text \cite{adelani2020generating}.
	This situation becomes even more serious when recent development of large pre-trained language models \cite{radford2019language, zellers2019defending} are capable of generating coherent, fluent and human-like text. Two influential works are GPT-2 \cite{radford2019language} and GROVER \cite{zellers2019defending}, The former is an open-sourced, large-scale unsupervised language model learned on web texts, while the latter is particularly learned for news. In this work, we study the problem of discriminating machine-generated and human-written text, and evaluate on datasets produced by both GPT-2 and GROVER.
	\par
	Advances in generative models have promoted the development of detection methods.
	Previous studies in the field of DeepFake detection of generated text are dominated by deep-learning based document classification models and studies about discriminating features of generated text. GROVER \cite{zellers2019defending} detects generated text by a fine-tuned model of the generative model itself. \citet{ippolito2019human} fine-tune the BERT model for discrimination and explore how sampling strategies and text excerpt length affect the detection. GLTR \cite{gehrmann2019gltr} develops a statistical method of computing per-token likelihoods and visualizes histograms over them to help deepfake detection. \citet{badaskar2008identifying}  and \citet{perez2017automatic} study language distributional features including n-gram frequencies, text coherence and syntax features. \citet{vijayaraghavan2020fake} study the effectiveness of different numeric representations (e.g., TFIDF and Word2Vec) and different neural networks (e.g., ANNs, LSTMs) for detection.
	\citet{bakhtin2019real} tackle the problem as a ranking task and study about the cross-architecture and cross-corpus generalization of their scoring functions. 
	\citet{schuster2019we} indicate that simple provenance-based detection methods are insufficient for solving the problem and call for development of fact checking systems.
	However, existing approaches struggle to capture fine-grained factual structures among continuous sentences, which in our observation is essential in discriminating human-written text and machine-generated text. Our approach takes a step towards modeling fine-grained factual structures for deepfake detection of text.
	
	\section{Conclusion}
	In this paper, we present \modelname, a graph-based reasoning approach utilizing fine-grained factual knowledge for DeepFake detection of text. 
	We represent the factual structure of a document as a graph, which is utilized to learn graph-enhanced sentence representations. Sentence representations are further composed through document-level aggregation for the final prediction, where the consistency and coherence of continuous sentences are sequentially modeled.
	We evaluate our system on a news-style dataset and a webtext-style dataset, whose fake instances are generated by GROVER and GPT-2 respectively. 
	Experiments show that components of our approach bring improvements and our full model significantly outperforms transformer-based baselines on both datasets. 
	 Model analysis further suggests that our model can distinguish the difference in the factual structure of machine-generated and human-written text.
	\section*{Acknowledgement}
	Wanjun Zhong, Zenan Xu, Jiahai Wang and Jian Yin are supported by the National Natural Science Foundation of China (U1711262, U1611264,U1711261,U1811261,U1811264, U1911203,62072483), National Key R\&D Program of China (2018YFB1004404), Guangdong Basic and Applied Basic Research Foundation (2019B1515130001),  Key R\&D Program of Guangdong Province (2018B010107005).  The corresponding author is Jian Yin.
	\bibliography{emnlp2020}
	\bibliographystyle{acl_natbib}
	\appendix
		\section{Training Details of NSP Model}
	\label{appendix:nsp}
	In this part, we describe the training details of our next sentence prediction model. 
	The training data of the NSP model comes from the human-written component of the webtext-style dataset or the news-style dataset depending on which dataset we are running experiments on. 
	We construct the dataset with balanced numbers of positive instances and negative instances. Supposing a positive instance is a continuous sentence pair ``$A;\ B$" from the human-written text, we construct a negative instance $``A; C"$, where $C$ is the most similar sentence in the document of $B$. 
	
	We tackle this problem as a binary classification task. We fine-tune the RoBERTa-Large model with standard cross-entropy loss on the binary classification task. We apply AdamW as the optimizer for model training. 
	We set the learning rate as 1e-5, batch size as 8, and set max sequence length as 128.
	
	\section{Training Details of FAST Model}
	\label{appendix:fast}
	In this part, we describe the
	training details for our experiments. We employ cross-entropy loss as the loss function. We apply AdamW as the optimizer for model training. We employ RoBERTa-Base as the backbone of our approach. 
	The RoBERTa network and graph-based reasoning model are trained jointly. 
	We set the learning rate as 1e-5, warmup step as 0, batch size as 4 per gpu, and set max sequence length as 512. 
	The training time for one epoch takes 2 hours on 4 P40 GPUs for the webtext-style dataset, and 20 minutes for the news-style dataset.
	We set the dimension of the contextual node representation as 100. The dimension of the wikipedia2vec entity representation is 100.
	
\end{document}